\documentclass[]{spie}  

\usepackage{amsmath,amsfonts,amssymb}
\usepackage{graphicx}
\usepackage[colorlinks=true, allcolors=blue]{hyperref}

\title{Text2Traffic: A Text-to-Image Generation and Editing Method for Traffic Scenes}

\newcommand{\cofirstauthor}{\textsuperscript{\dag}}
\newcommand{\correspondingauthor}{\textsuperscript{*}}

\author[a]{Feng Lv\cofirstauthor}
\author[a]{Haoxuan Feng\cofirstauthor}
\author[a]{Zilu Zhang}
\author[a]{Chunlong Xia\correspondingauthor}
\author[b]{Yanfeng Li\correspondingauthor}
\affil[a]{Baidu, Beijing, China}
\affil[b]{School of Electronic and Information Engineering, Beijing Jiaotong University, Beijing, China}

\authorinfo{\\E-mail: \{lvfeng02, fenghaoxuan, zhangzilu01, xiachunlong\}@baidu.com, yf.li@bjtu.edu.cn \\  \dag  represent equal contribution, * represent corresponding authors}

\begin{document} 
\maketitle

\begin{abstract}
With the rapid advancement of intelligent transportation systems, text-driven image generation and editing techniques have demonstrated significant potential in providing rich, controllable visual scene data for applications such as traffic monitoring and autonomous driving. However, several challenges remain, including insufficient semantic richness of generated traffic elements, limited camera viewpoints, low visual fidelity of synthesized images, and poor alignment between textual descriptions and generated content. To address these issues, we propose a unified text-driven framework for both image generation and editing, leveraging a controllable mask mechanism to seamlessly integrate the two tasks. Furthermore, we incorporate both vehicle-side and roadside multi-view data to enhance the geometric diversity of traffic scenes. Our training strategy follows a two-stage paradigm: first, we perform conceptual learning using large-scale coarse-grained text-image data; then, we fine-tune with fine-grained descriptive data to enhance text-image alignment and detail quality. Additionally, we introduce a mask-region-weighted loss that dynamically emphasizes small yet critical regions during training, thereby substantially enhancing the generation fidelity of small-scale traffic elements. Extensive experiments demonstrate that our method achieves leading performance in text-based image generation and editing within traffic scenes.  
\end{abstract}


\section{INTRODUCTION}
\label{sec:intro}  

In recent years, Diffusion Models have driven the rapid advancement of Text-to-Image Generation and Image Editing technologies, thanks to their exceptional generative capabilities and ability to model complex data distributions.~\cite{ho2020denoising} These technologies are not only widely applied in fields such as artistic creation, virtual reality, and content generation but are also beginning to demonstrate potential in specialized vertical domains—particularly in Intelligent Transportation Systems. By generating or editing traffic scene images based on natural language instructions, these technologies can provide rich, controllable, and diverse visual data for tasks such as traffic monitoring, autonomous driving simulation, and road safety analysis. This, in turn, helps alleviate bottlenecks associated with the high cost of real data collection, difficulties in annotation, and limited scene coverage.

However, directly transferring general text-to-image models to traffic scenes faces numerous challenges. Firstly, existing models often lack semantic richness and structural coherence when generating traffic-related elements, making it difficult to meet the detail accuracy required for professional applications. Secondly, most approaches support image generation from a single perspective (either vehicle-side or roadside), lacking joint modeling of multi-view data, which limits the geometric consistency and semantic completeness of the scenes. Additionally, the fidelity of generated images is relatively low, particularly in the representation of small objects like trafficcone, where details may appear blurred or distorted. More crucially, the semantic alignment between text descriptions and generated images is often insufficiently precise in complex traffic semantics, leading to a divergence between the generated content and user intent.

To tackle the aforementioned challenges, we propose a unified text-driven framework specifically designed for image generation and editing in traffic scenes. Our core insight is to unify generation and editing tasks through the use of controllable mask regions, allowing for precise manipulation of specific traffic elements. Building on this approach, we combine both vehicle-side and roadside perspectives to create a training dataset that offers enhanced geometric diversity and semantic richness. For the training strategy, we employ a two-stage learning paradigm to significantly enhance text-image alignment and detail fidelity. Additionally, to address the challenges of generating small-scale objects, we design a mask-region-weighted loss function that assigns greater learning weights to critical regions during training. This strengthens the model's ability to accurately represent small yet semantically important traffic elements.

The main contributions of this work are summarized as follows:
(1) We present the first unified text-driven framework for image generation and editing specifically designed for traffic scenes, supporting both text-to-image synthesis and precise image editing.
(2) We construct a multi-view traffic text-image dataset that fuses vehicle-side and roadside perspectives, and propose a two-stage training strategy that effectively enhances the semantic accuracy and visual fidelity of generated content.
(3) We introduce a mask-region-weighted loss function that significantly improves the generation quality of small-scale traffic objects.
(4) Extensive experiments demonstrate the superiority of our method over existing approaches in traffic-oriented image generation and editing tasks, offering a novel and practical pathway for advancing intelligent transportation systems.

\begin{figure*}[t]
\begin{center}
\includegraphics[width = 1.0 \linewidth]{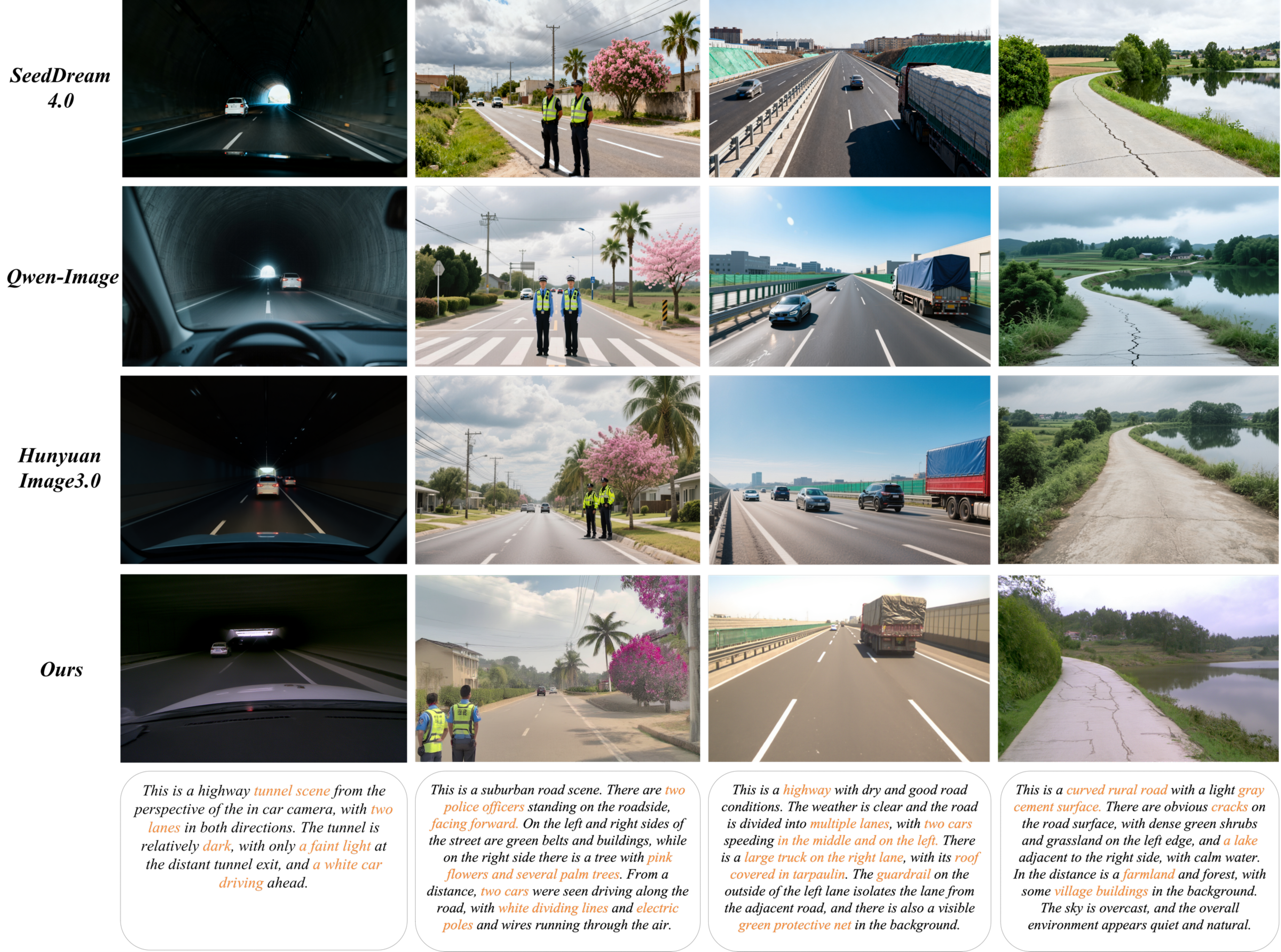}
\end{center}
\vspace{-2mm}
\caption[example] 
{ \label{car_t2i} 
Comparison of text-to-image performance for vehicle-side cameras across different models (key descriptions highlighted in yellow)}
\end{figure*}

\section{Related Work}
\label{sec:related}
\subsection{Diffusion Models}
In the field of text-to-image generation and image editing, diffusion models have become the mainstream paradigm. Stable Diffusion(SD)~\cite{rombach2022high}  effectively alleviates the computational bottleneck encountered in training within high-dimensional pixel spaces by transferring the diffusion process to a latent space and introducing text-conditioned control, while achieving high-quality text-to-image generation. The subsequent SDXL~\cite{podell2023sdxl} adopts a "Base + Refiner" two-stage architecture, where the Base model generates a low-resolution image, and the Refiner model enhances details in the high-resolution latent space, significantly improving the fidelity and semantic consistency of the generated images. To further enhance spatial control over the generation process, ControlNet~\cite{zhang2023adding} was proposed. By introducing pluggable auxiliary networks, it supports precise guidance from various spatial conditions, such as edge maps, depth maps, and pose maps without compromising the original pre-trained model, greatly expanding the application boundaries of image editing and controlled generation.

Recently, Diffusion Transformer(DiT)~\cite{peebles2023scalable} has integrated the transformer architecture into diffusion models, replacing the convolutional structures in traditional U-Nets with self-attention mechanisms to better model long-range dependencies and global semantic relationships. DiT has great scalability and generation quality, making it the basis for the next generation of text-to-image models. Building on this foundation, models such as FLUX~\cite{flux}, SD3~\cite{esser2024scaling}, Lumina Image 2.0~\cite{qin2025lumina}, HunyuanImage 3.0~\cite{cao2025hunyuanimage}, SeedDream 4.0~\cite{seedream2025seedream} and Qwen-Image~\cite{wu2025qwen} have expanded based on the DiT architecture, achieving significant advancements in generation quality, multilingual understanding, and multimodal alignment, progressively moving towards a unified multimodal generation representation. However, their effectiveness in text to image and image editing tasks in traffic scenes is not commendable.

Concurrently, multimodal large language models (MLLMs) like Qwen2.5-VL~\cite{bai2025qwen2} and InternVL3~\cite{zhu2025internvl3} are widely used for the automatic generation of high-quality image descriptions. These models can generate semantically rich and fine-grained textual annotations, providing more precise supervision signals for text-to-image models, further enhancing the coherence and diversity between generated images and text prompts.


\begin{figure*}[t]
\begin{center}
\includegraphics[width=1.0\linewidth]{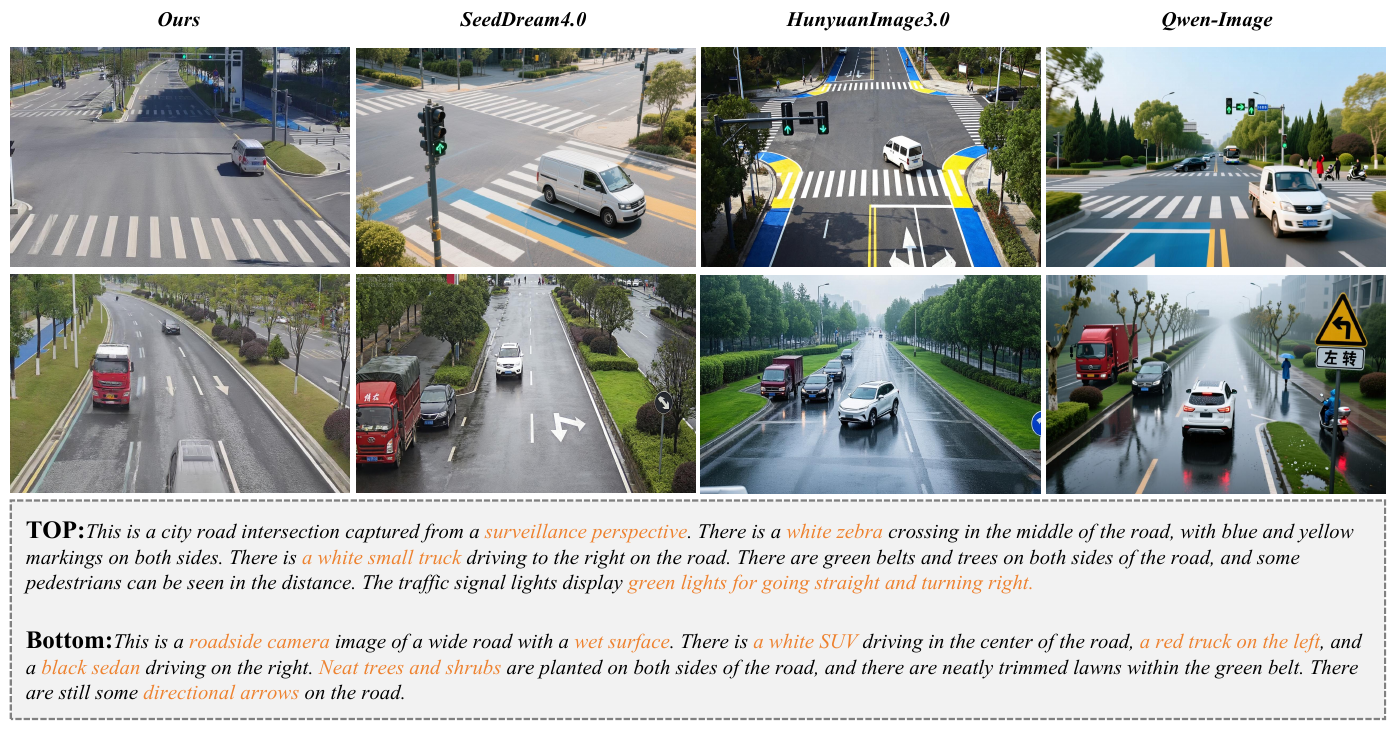}
\end{center}
\vspace{-2mm}
\caption[example] 
{ \label{luce_t2i} 
Comparison of text-to-image performance for roadside cameras across different models.}
\end{figure*}

\subsection{Image Generation and Editing in Traffic Scenes}
In recent years, text-to-image generation and image editing for traffic scenes have gained increasing attention. Text2Street~\cite{gu2025text2street}, for instance, utilizes a lane-aware road topology generator to control road layouts in a text-to-map manner, enabling the creation of street scene images from textual descriptions. BEVGen~\cite{swerdlow2024street} combines an innovative cross-view transformation with spatial attention design, learning the relationship between camera and map views to synthesize realistic and spatially consistent surrounding images that align with the bird's-eye view (BEV) layout of traffic scenes. DriveDreamer~\cite{wang2024drivedreamer} focuses on generating high-quality, controllable driving videos that reflect real-world traffic structures. In the realm of image editing, LightDiff~\cite{chen2024lightdiff} uses a multi-condition controlled diffusion model, incorporating depth maps, RGB images, and text captions to effectively illuminate dark scenes, aiming to enhance low-light image quality for autonomous driving applications. However, despite these approaches introducing various generation conditions, they are unable to support text-based multi-view traffic scene image generation and editing, thereby limiting the generation of specific scene data. In this paper, we integrate image generation and editing from both vehicle-side and roadside camera perspectives in traffic scenes. This method supports the creation of a wider range of traffic elements and scenes, enhancing their diversity and fidelity. As a result, it provides more valuable data for applications such as autonomous driving, V2X, and beyond.

\section{Method}
\label{sec:method}
\subsection{A Multi-Level Captioning Framework}
The performance of vision-language models is critically dependent on the quality and richness of their training data, with high-quality image captions playing a pivotal role in achieving state-of-the-art results. Consequently, the use of pre-trained Vision Language Models (VLMs) for image captioning has become a standard practice in recent literature. However, these models suffer from significant limitations, including single-granularity descriptions, domain bias, and constraints from fixed, low-resolution input sizes. These shortcomings often result in suboptimal caption quality that fails to align with the complexity and specificity of real-world user prompts, creating a large performance gap.

\begin{figure*}[t!]
\begin{center}
\includegraphics[width=1.0\linewidth]{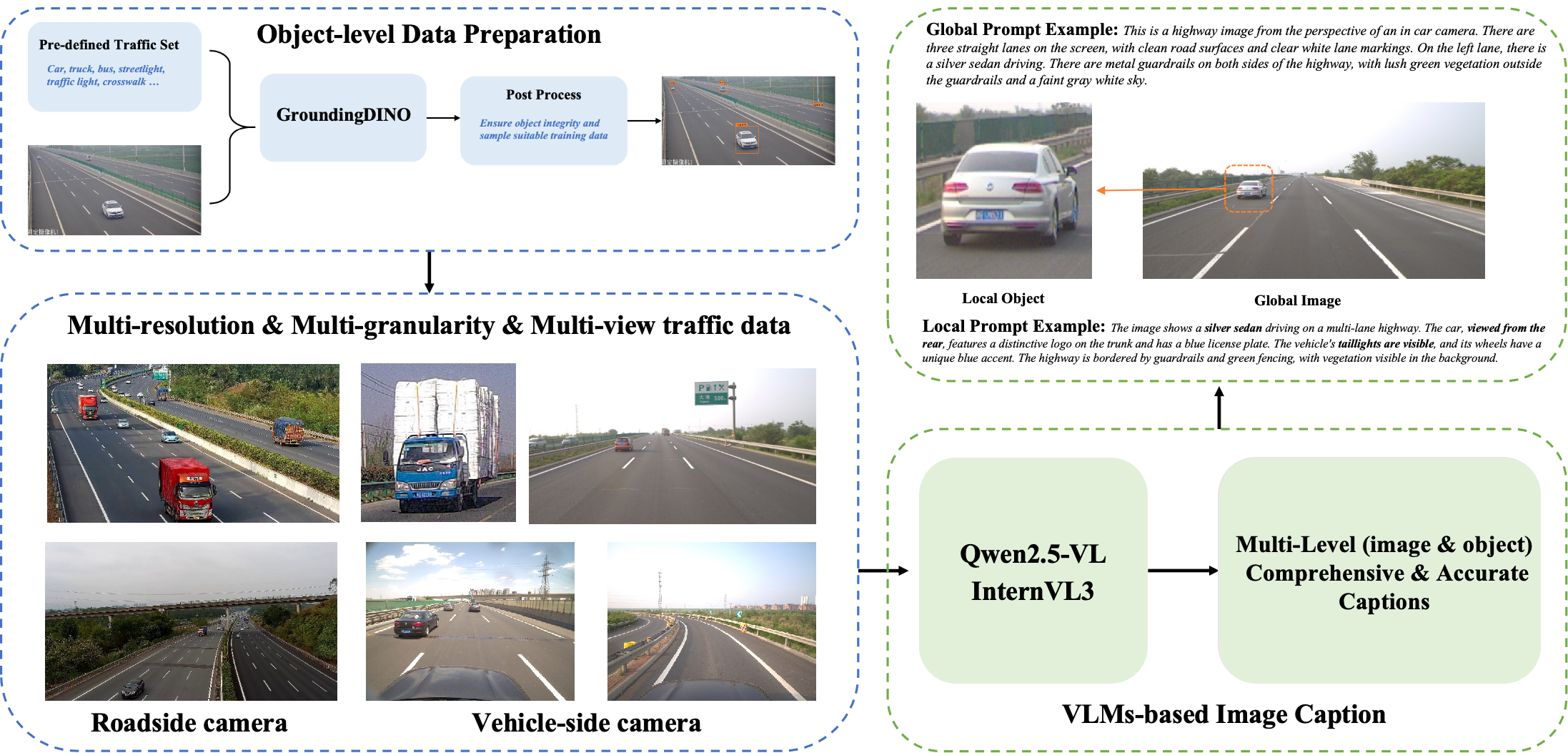}
\end{center}
\vspace{-1.5mm}
\caption[example]
{Process of image description generation method.}
\label{Caption}
\end{figure*}

To address these challenges and construct a well-aligned text-to-image traffic dataset, we developed a refined data preparation pipeline shown in Fig.~\ref{Caption}. This process aims to generate high-quality, multi-modal annotations necessary for robust model training in traffic domains. Specifically, the pipeline begins with the acquisition of multi-resolution and multi-view traffic images, captured from both roadside and vehicle-side perspectives to ensure holistic geometric consistency and semantic completeness of the scene. A fundamental and distinctive step in our methodology involves processing this domain-specific data with an open-set detection model, such as GroundingDINO~\cite{liu2024grounding}. Following the initial detection, we apply Non-Maximum Suppression (NMS) to eliminate redundant and overlapping bounding boxes, ensuring that each detected object is uniquely and accurately localized. Furthermore, to enhance the quality and reliability of the detected objects, we implement a size-based filtering strategy that removes exceptionally small objects lacking distinctive features. This filtering process is essential for maintaining object integrity and preventing the inclusion of ambiguous or noisy visual elements that could hinder model performance.

After the open-set detection process, the pipeline uses advanced VLMs, such as Qwen2.5-VL 7B and InternVL3 8B, to implement a multi-level captioning strategy. This strategy produces two complementary types of descriptions: 1) a global, image-level caption that captures the overall scene context to ensure scene consistency; and 2) local, object-level captions that offer detailed attributes for each detected traffic entity. This hierarchical annotation approach effectively meets the diverse needs of text-to-image generation, which requires both a comprehensive understanding of the scene and precise object reference and manipulation for image editing.

\begin{figure*}[t]
\begin{center}
\includegraphics[width=1.0\linewidth]{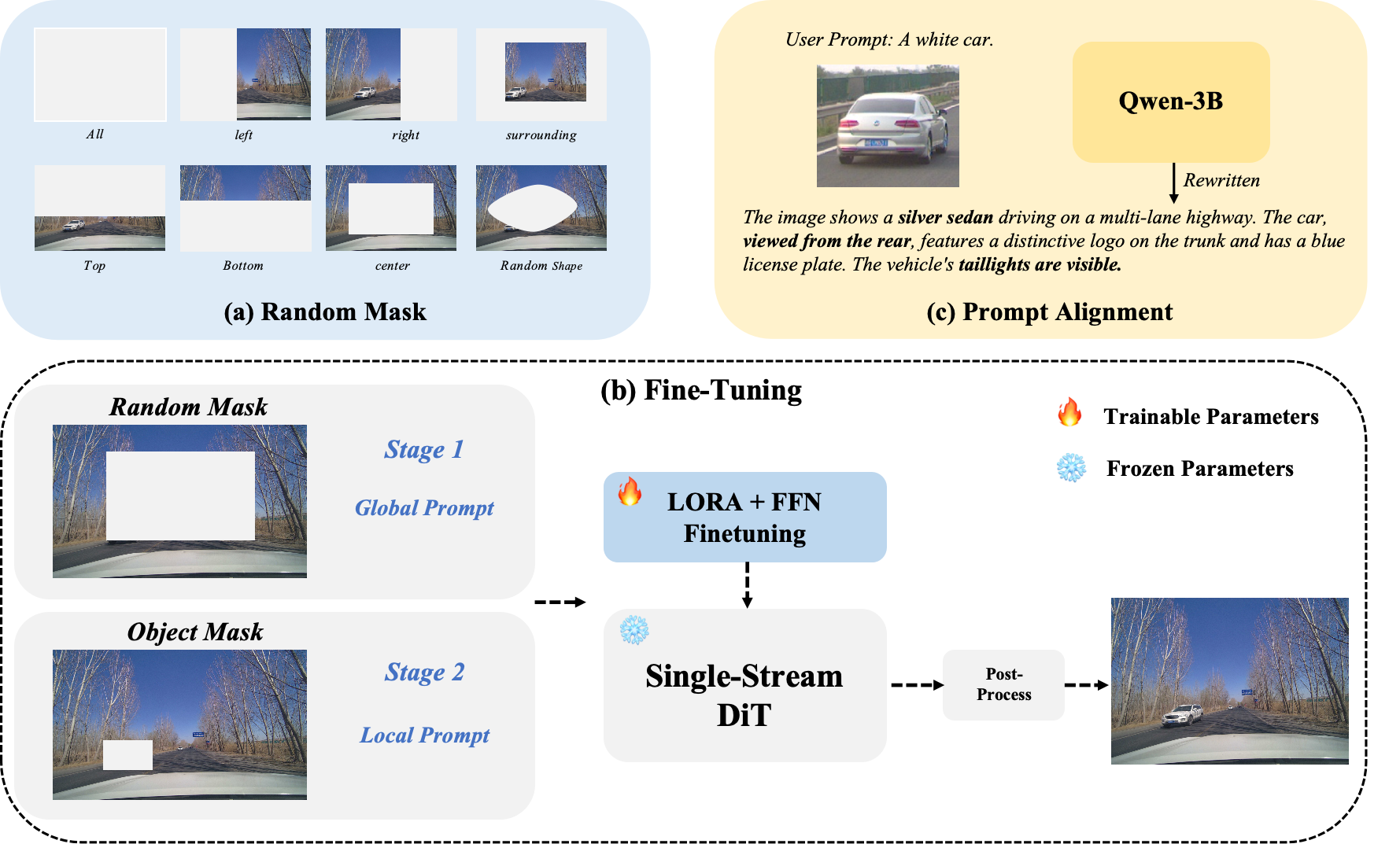}
\end{center}
\caption[example] 
{Our restoration-based supervised fine-tuning method.}
\label{finetune}
\end{figure*}

\subsection{Restoration-based Supervised Fine-Tuning}
A significant challenge in vision-language modeling is integrating the seemingly distinct tasks of text-to-image generation and semantic image editing into a single, cohesive framework. To address this, we conceptualize both tasks as a unified image restoration problem. Given an input image \( I \) with a mask \( m \), a deep learning network parameterized by \( f_\theta \), and some guidance \( c \), the process can be represented as \( v = f_\theta(I \cdot m, c) \). Specifically, in our proposed Restoration-based Supervised Fine-Tuning (R-SFT) approach, the model receives a masked image and a conditioning text caption. Its objective is to predict the velocity field that facilitates the restoration of the original, uncorrupted image, effectively learning to "inpaint" the missing regions in a semantically accurate manner based on the textual guidance.

\begin{figure*}[t]
\begin{center}
\includegraphics[width=0.85\linewidth]{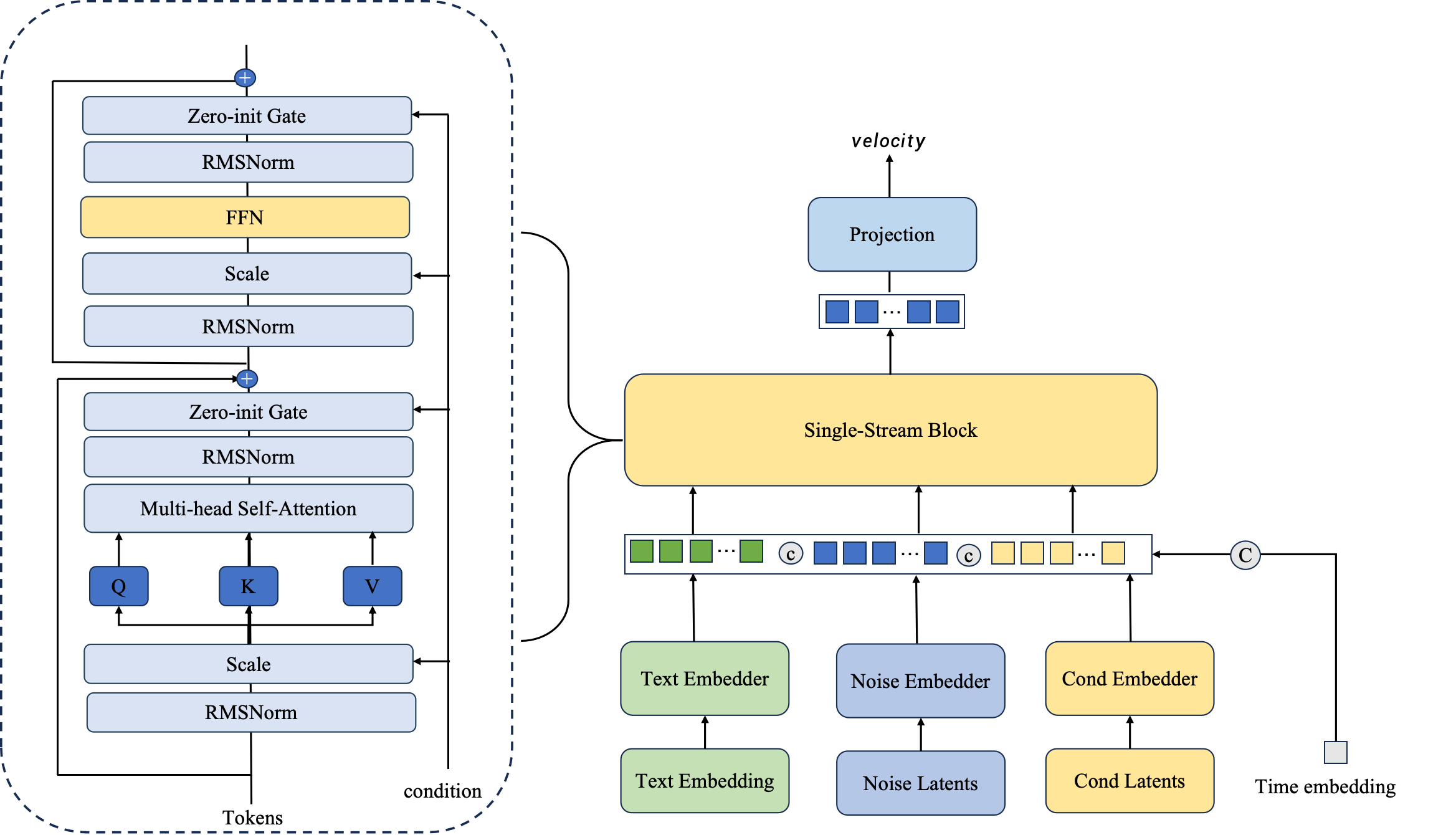}
\end{center}
\vspace{-1mm}
\caption[example] 
{Model architecture details.}
\label{model}
\end{figure*}

Fig.~\ref{finetune} illustrates the architecture of our proposed R-SFT framework, a novel training paradigm designed to unify and enhance image generation and editing capabilities within a single model. The framework is structured into three core, interconnected components: (a) Random Mask, (b) Fine-Tuning, and (c) Prompt Alignment, which collectively guide the model from data preparation to high-fidelity output generation. The process initiates with the random mask module (a), which preprocesses the original training images by applying variable-shaped and sized occlusions. This deliberate corruption transforms the task into the restoration problem, forcing the model to learn robust contextual reasoning and inpainting skills essential for both conditional generation and editing. The fine-tuning component (b) forms the core of the training pipeline, implemented as a two-stage progressive learning strategy. In Stage 1, the model is trained to reconstruct images from large-area masks, focusing on recovering broad scene context and global consistency. Stage 2 refines these capabilities using smaller, object-level masks, honing the model's ability to perform detailed, precise restoration. Both stages leverage a single-stream diffusion transformer ~\cite{qin2025lumina} as a backbone, which is efficiently optimized using a combination of LoRA~\cite{hu2022lora} and FFN fine-tuning. The details of the DiT architecture are shown in Fig.~\ref{model}. In the context of diffusion transformer models, a notable advantage of the single-stream architecture lies in its capacity to seamlessly incorporate supplementary conditional inputs. Unlike alternative approaches that necessitate the introduction of additional module branches, such as adapters or ControlNet~\cite{zhang2023adding}, the single-stream DiT can directly integrate conditional information without the need for such structural modifications. To leverage this characteristic for the inpainting task, we introduce a dedicated conditional embedder. This embedder is specifically designed to process the inpainting image, denoted as $I_m$, obtained by element-wise multiplication of the original image $I$ and the mask $m$. The output of the conditional embedder is then concatenated with noisy latent representations and text tokens. Subsequently, a self-attention mechanism is employed to facilitate multi-modal interaction among these concatenated features. This approach enables the model to effectively combine visual information from the inpainting region, latent representations of the corrupted image, and semantic information from the text prompt, thereby enhancing the overall inpainting performance.

Crucially, the Prompt Alignment module (c) operates during inference to ensure semantic consistency. It utilizes a powerful pretrained Large-Language Model (Qwen-3B~\cite{bai2023qwen}) to interpret the user's textual prompt and align it to the caption distribution. This mechanism provides explicit semantic guidance, enabling the model to perform prompt-aware restoration and object edits. By integrating a restoration-based objective with a structured fine-tuning schedule and a dedicated prompt-alignment mechanism, the R-SFT framework effectively equips a unified model with powerful and versatile image manipulation capabilities.

\subsection{Training objective}
We employ flow matching to optimize the parameters of our model. In contrast to Denoising Diffusion Probabilistic Models (DDPM)~\cite{ho2020denoising}, flow matching performs the forward process by linearly interpolating between noise and data along a straight trajectory. At any given timestep $t$, the noisy sample $x_t$ is defined as: $x_t = t \cdot x + (1 - t) \cdot \epsilon$, where x represents the original data and $\epsilon \sim \mathcal{N}(0, I)$ denotes Gaussian noise. The model is trained to directly regress the target velocity vector $v$, conditioned on the noisy input $x_t$, the timestep $t$, and any auxiliary conditioning information $c$. The flow matching loss:
\begin{equation}
    L_{flow} (\phi) = E[||v_\phi (x_t, t, c) - (x - \epsilon)||^2]
\end{equation}

Additionally, we introduce a region-weighted loss to direct the model's focus towards the masked regions that require inpainting or editing. This is a critical need for generating small, distinct objects like cones and pedestrians in traffic scenes. Diffusion models often fail to consider fine-grained details in localized edits. This can result in blurry or inconsistent generation in small areas. This phenomenon arises from the standard mean-squared error loss function, which treats all pixels in an identical manner. Consequently, the gradient signal from a small edited region is overwhelmed by the substantial unchanged background. The weighted loss as follows:
\begin{equation}
    w_{i,j} = \left\{
    \begin{array}{ll}
    1 & \text{if } \mathbf{I}_{i,j} = \mathbf{I}_{i,j}^{\prime} \\
    \frac{1}{\|\mathbf{I} - \mathbf{I}^{\prime}\|^{2}} & \text{if } \mathbf{I}_{i,j} \neq \mathbf{I}_{i,j}^{\prime}
    \end{array}
    \right.
\end{equation}
Where $I, I^{\prime}$ represent the latent representations of input image and target image.  This loss function strategically assigns higher weights to the edited areas, ensure the model to allocate better representational capability to generating and refining the critical parts. The combination of flow matching loss and the proposed weighted loss ensures that the model not only achieves high-fidelity restoration but also exhibits heightened sensitivity and accuracy in editing regions, effectively addressing the challenge of small-object generation.

\section{Experiment}
\label{}

\subsection{Experiment Details}
Our dataset consists of three components: open-source datasets (BDD100K~\cite{yu2020bdd100k}, TT100K~\cite{zhu2016traffic}, CUHK-SYSU~\cite{xiao2017joint}), an internal proprietary dataset, and an internet dataset. The training set comprises approximately 2.4M text-image pairs, including 2M pairs with detailed global image descriptions and 0.4M pairs with local object-level descriptions. The validation set is carefully curated, consisting of 100 samples. This data encompasses a wide variety of scenes in the traffic domain, including different perspectives, elements, weather conditions, and lighting variations. Our model is trained on 8xA800 GPUs using the AdamW~\cite{loshchilov2017decoupled} optimizer, with an initial learning rate of 1e-4 and a global batch size of 32. The training supports a dynamic resolution strategy and is conducted in two stages. In the first stage, we train for 16 epochs using global image descriptions to capture the high-level semantic structure of traffic scenes. In the second stage, we fine-tune for 5 epochs, using local object descriptions to refine details and enhance local realism. Evaluation metrics such as FID~\cite{heusel2017gans}, SSIM, and PSNR are used to assess the model's convergence and generation quality.

\subsection{Qualitative Analysis}
To comprehensively evaluate our model's generative capabilities in real-world traffic scenarios, we conducted an in-depth qualitative comparison from two key perspectives: vehicle-side  and roadside camera. We compared our model with the current state-of-the-art open-source generative models, including Qwen-Image, SeedDream 4.0, iRAG, and HunyuanImage 3.0. Our evaluation focused on two major tasks: text-based image editing and text-to-image generation. We specifically examined the models' performance on core metrics such as geometric consistency, precise adherence to instructions, text-image semantic alignment, and hallucination suppression.


\begin{figure}[t!]
\begin{center}
\includegraphics[width = 1.0\linewidth]{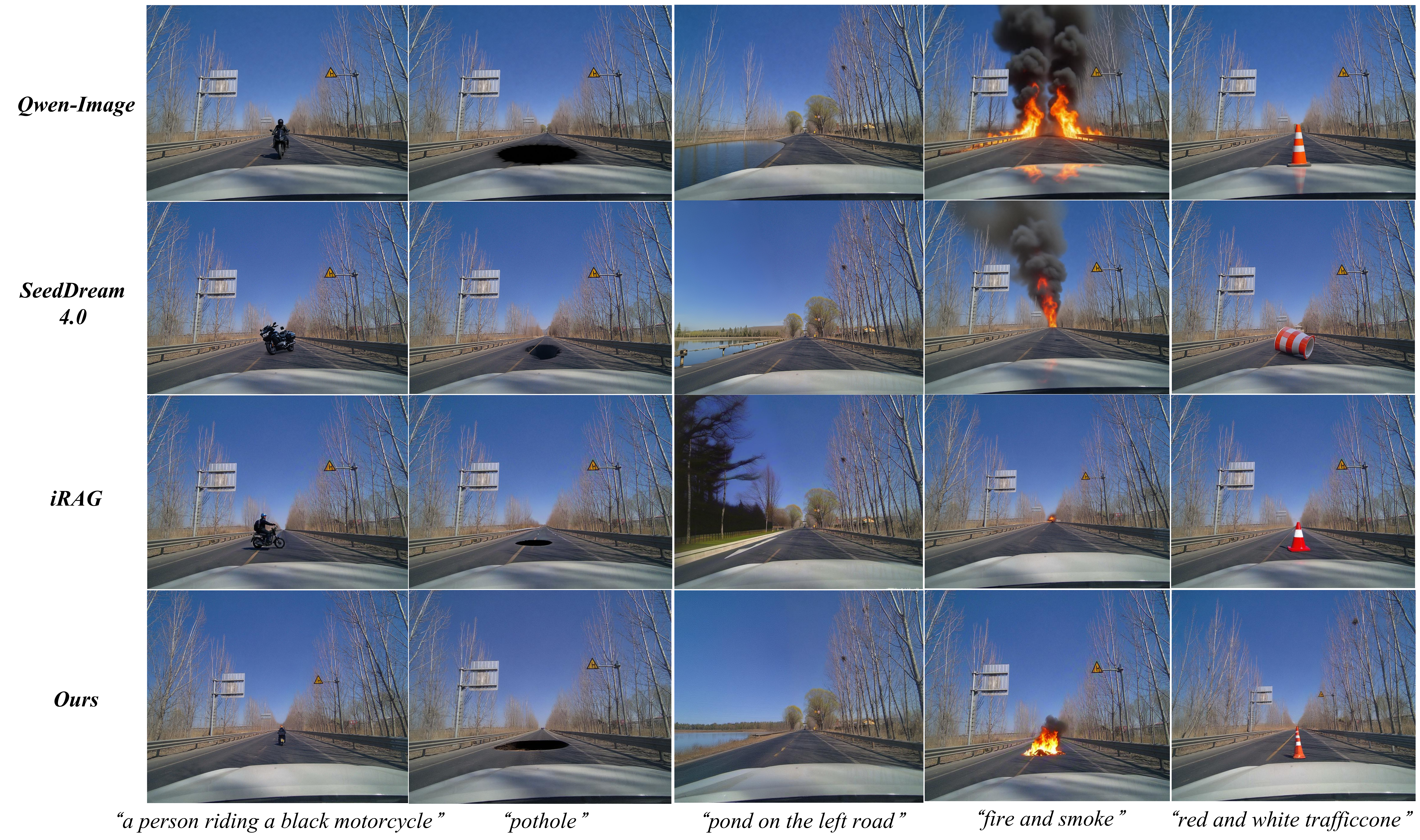}
\end{center}
\vspace{-2mm}
\caption[example] 
{ \label{car_inpainting} 
Comparison of text-based image editing performance for vehicle-side cameras across different models.}
\end{figure}

\subsubsection{Vehicle-side perspective}
As shown in Fig.~\ref{car_inpainting}, we compared different models' performance in editing a vehicle-side image based on the same textual instructions. Our model excels in generating semantically and geometrically coherent modifications, seamlessly incorporating elements like "road potholes," "traffic cones," and a "pond along the left roadside" while maintaining the scene's perspective and spatial structure. In contrast, models like Qwen-Image and SeedDream 4.0 often produce objects that disrupt spatial consistency, showing scale distortions and misalignment with the scene's 3D geometry.

Fig.~\ref{car_t2i} compares the text-to-image generation capabilities of different models across various scenarios. In challenging environments such as dark tunnels and rural roads, all models accurately depicted the respective scenes, but our generated images exhibited higher fidelity. In suburban and highway scenarios, Qwen-Image and HunyuanImage3.0 experienced issues with hallucinations and content inconsistency. Our model, however, adhered strictly to detailed instructions such as "face forward" and "two cars in the middle and left road," achieving superior text-image semantic alignment and instruction following.


\begin{figure*}[h]
\begin{center}
\includegraphics[width = 1.0\linewidth]{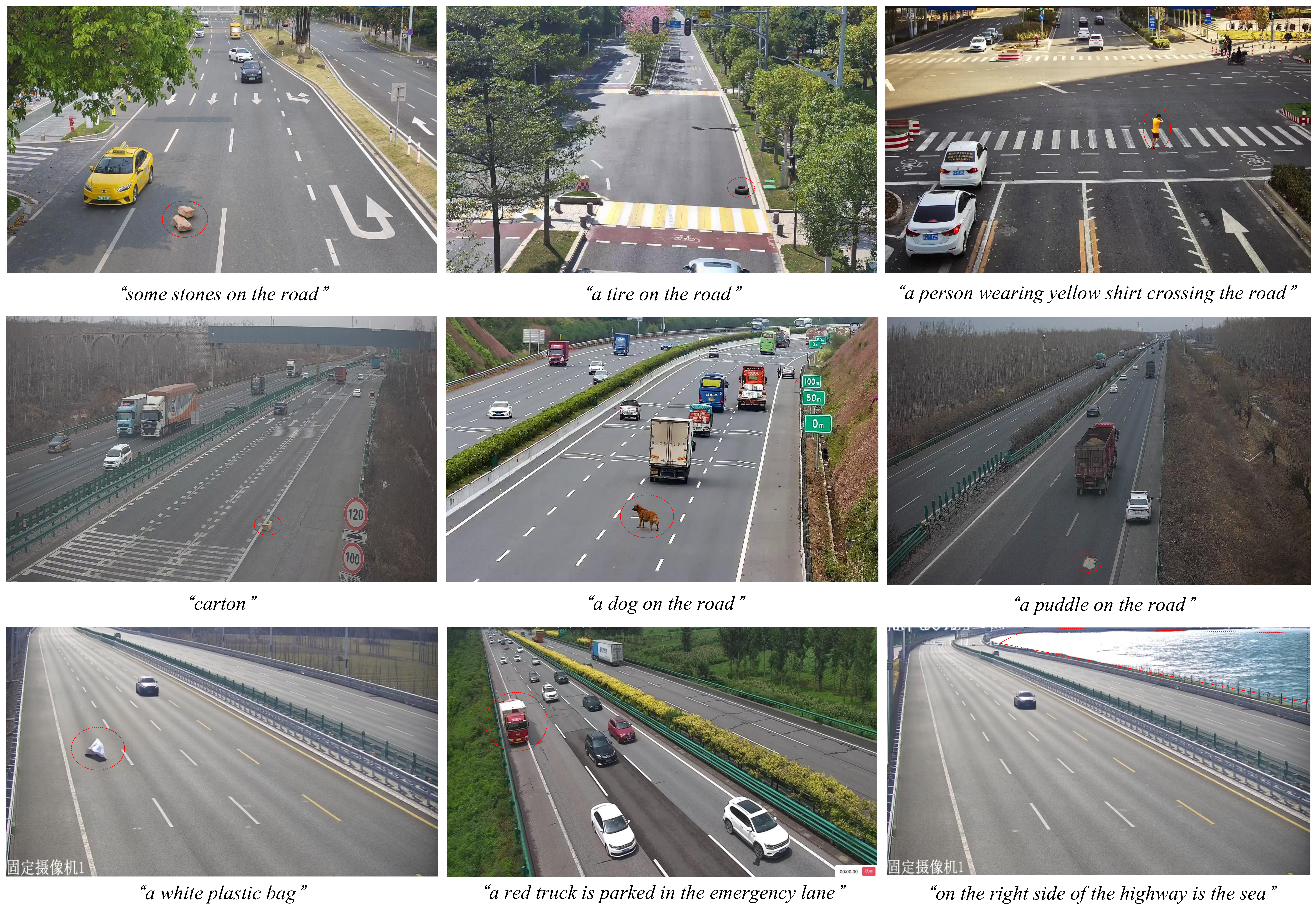}

\end{center}
\vspace{-1mm}
\caption[example] 
{ \label{luce_inpaint} 
Image editing results for multiple rare traffic elements from a roadside cameras, encompassing both urban and highway scenes.}
\end{figure*} 

\subsubsection{Roadside perspective}
Fig.~\ref{luce_inpaint} presents the results of our model editing various rare traffic elements from a roadside perspective, such as "stones on the road," "cardboard boxes," "dogs," "puddles," and "white plastic bags." Our model consistently and accurately integrates these elements into complex urban and highway scenes, ensuring that the size, position, lighting, and shadows of the objects are highly consistent with the original image. Due to the broader coverage of the roadside perspective, the model requires a higher level of global understanding and detail control. This demonstrates the model's strong geometric consistency and text-image alignment capabilities.

Fig.~\ref{luce_t2i} showcases the text-to-image generation performance of different models. Notably, Qwen-Image does not support image generation under this perspective. Given the increased number of traffic elements and the more diverse and complex nature of roadside scenes, all models exhibit some challenges in achieving precise text-image alignment. Addressing these challenges will be a key focus for future optimization efforts.

\begin{table}[h]
\setlength{\tabcolsep}{12.0 pt} 
\renewcommand{\arraystretch}{1.2}
\caption{Comparison of metrics at different training stages.} 
\label{tab:exp}
\begin{center}  
\begin{tabular}{l|c|c|c}
\hline
       & FID & SSIM & PSNR  \\ \hline
$\text{Stage1}$ & 9.8 & 0.77 & 21.06 \\ \hline
$ \text{Stage2}$ & 9.7 & 0.79 & 24.13 \\ \hline
$ + \text{Weighted Loss}$ & 9.6 & 0.81 & 24.52 \\ \hline
\end{tabular}
\end{center}
\end{table}

\subsection{Quantative Analysis}
We evaluated our two-stage model on the validation set, as shown in Tab.~\ref{tab:exp}. In Stage 1, we pre-trained the model using large-scale image-text pairs, achieving FID = 9.8, SSIM = 0.77, and PSNR = 21.06. In Stage 2, we fine-tuned the model with fine-grained object descriptions, resulting in improved performance: FID decreased to 9.7, while SSIM and PSNR increased to 0.79 and 24.13, respectively. An ablation study confirmed that our masked region-weighted loss function decreased FID further, from 9.7 to 9.6, while PSNR and SSIM increased to 0.81 and 24.52 respectively. This demonstrates that focusing loss weighting on object areas improves both detail fidelity and semantic accuracy.

\section{Conclusion}
In this paper, we propose Text2Traffic, a unified text-driven framework for both image generation and editing in traffic scenes. Our method addresses key challenges in this domain, including limited semantic richness, restricted camera viewpoints, low visual fidelity, and poor text-image alignment. Also, the two-stage training strategy—combining large-scale conceptual learning with fine-grained descriptive fine-tuning—effectively improves semantic alignment and detail quality. The introduced mask-region-weighted loss function further boosts the fidelity of small-scale traffic elements, which are critical for practical applications. Extensive experiments validate that our method achieves superior performance in both generation and editing tasks, offering a robust solution for creating high-quality, controllable visual data to advance intelligent transportation systems.


\acknowledgments 
The work was supported in part by the National Natural Science Foundation of China (62272027),and Beijing Natural Science Foundation (4232012, L252076).

\bibliography{main} 

@article{bai2025qwen2,
  title={Qwen2. 5-vl technical report},
  author={Bai, Shuai and Chen, Keqin and Liu, Xuejing and Wang, Jialin and Ge, Wenbin and Song, Sibo and Dang, Kai and Wang, Peng and Wang, Shijie and Tang, Jun and others},
  journal={arXiv preprint arXiv:2502.13923},
  year={2025}
}

@article{zhu2025internvl3,
  title={Internvl3: Exploring advanced training and test-time recipes for open-source multimodal models},
  author={Zhu, Jinguo and Wang, Weiyun and Chen, Zhe and Liu, Zhaoyang and Ye, Shenglong and Gu, Lixin and Tian, Hao and Duan, Yuchen and Su, Weijie and Shao, Jie and others},
  journal={arXiv preprint arXiv:2504.10479},
  year={2025}
}

@article{seedream2025seedream,
  title={Seedream 4.0: Toward next-generation multimodal image generation},
  author={Seedream, Team and Chen, Yunpeng and Gao, Yu and Gong, Lixue and Guo, Meng and Guo, Qiushan and Guo, Zhiyao and Hou, Xiaoxia and Huang, Weilin and Huang, Yixuan and others},
  journal={arXiv preprint arXiv:2509.20427},
  year={2025}
}

@article{podell2023sdxl,
  title={Sdxl: Improving latent diffusion models for high-resolution image synthesis},
  author={Podell, Dustin and English, Zion and Lacey, Kyle and Blattmann, Andreas and Dockhorn, Tim and M{\"u}ller, Jonas and Penna, Joe and Rombach, Robin},
  journal={arXiv preprint arXiv:2307.01952},
  year={2023}
}

@inproceedings{peebles2023scalable,
  title={Scalable diffusion models with transformers},
  author={Peebles, William and Xie, Saining},
  booktitle={Proceedings of the IEEE/CVF international conference on computer vision},
  pages={4195--4205},
  year={2023}
}

@inproceedings{rombach2022high,
  title={High-resolution image synthesis with latent diffusion models},
  author={Rombach, Robin and Blattmann, Andreas and Lorenz, Dominik and Esser, Patrick and Ommer, Bj{\"o}rn},
  booktitle={Proceedings of the IEEE/CVF conference on computer vision and pattern recognition},
  pages={10684--10695},
  year={2022}
}

@article{cao2025hunyuanimage,
  title={Hunyuanimage 3.0 technical report},
  author={Cao, Siyu and Chen, Hangting and Chen, Peng and Cheng, Yiji and Cui, Yutao and Deng, Xinchi and Dong, Ying and Gong, Kipper and Gu, Tianpeng and Gu, Xiusen and others},
  journal={arXiv preprint arXiv:2509.23951},
  year={2025}
}

@article{qin2025lumina,
  title={Lumina-image 2.0: A unified and efficient image generative framework},
  author={Qin, Qi and Zhuo, Le and Xin, Yi and Du, Ruoyi and Li, Zhen and Fu, Bin and Lu, Yiting and Yuan, Jiakang and Li, Xinyue and Liu, Dongyang and others},
  journal={arXiv preprint arXiv:2503.21758},
  year={2025}
}

@inproceedings{zhang2023adding,
  title={Adding conditional control to text-to-image diffusion models},
  author={Zhang, Lvmin and Rao, Anyi and Agrawala, Maneesh},
  booktitle={Proceedings of the IEEE/CVF international conference on computer vision},
  pages={3836--3847},
  year={2023}
}

@inproceedings{gu2025text2street,
  title={Text2street: Controllable text-to-image generation for street views},
  author={Gu, Songen and Su, Jinming and Duan, Yiting and Chen, Xingyue and Luo, Junfeng and Zhao, Hao},
  booktitle={International Conference on Pattern Recognition},
  pages={130--145},
  year={2025},
  organization={Springer}
}

@article{swerdlow2024street,
  title={Street-view image generation from a bird's-eye view layout},
  author={Swerdlow, Alexander and Xu, Runsheng and Zhou, Bolei},
  journal={IEEE Robotics and Automation Letters},
  volume={9},
  number={4},
  pages={3578--3585},
  year={2024},
  publisher={IEEE}
}

@inproceedings{wang2024drivedreamer,
  title={Drivedreamer: Towards real-world-drive world models for autonomous driving},
  author={Wang, Xiaofeng and Zhu, Zheng and Huang, Guan and Chen, Xinze and Zhu, Jiagang and Lu, Jiwen},
  booktitle={European conference on computer vision},
  pages={55--72},
  year={2024},
  organization={Springer}
}

@inproceedings{chen2024lightdiff,
  title={LighTDiff: surgical endoscopic image low-light enhancement with T-diffusion},
  author={Chen, Tong and Lyu, Qingcheng and Bai, Long and Guo, Erjian and Gao, Huxin and Yang, Xiaoxiao and Ren, Hongliang and Zhou, Luping},
  booktitle={International Conference on Medical Image Computing and Computer-Assisted Intervention},
  pages={369--379},
  year={2024},
  organization={Springer}
}

@inproceedings{liu2024grounding,
  title={Grounding dino: Marrying dino with grounded pre-training for open-set object detection},
  author={Liu, Shilong and Zeng, Zhaoyang and Ren, Tianhe and Li, Feng and Zhang, Hao and Yang, Jie and Jiang, Qing and Li, Chunyuan and Yang, Jianwei and Su, Hang and others},
  booktitle={European conference on computer vision},
  pages={38--55},
  year={2024},
  organization={Springer}
}

@inproceedings{flux,
title={Black Forest Labs. Flux. https://github.com/black-forest-labs/flux, 2023.}
}

@inproceedings{esser2024scaling,
  title={Scaling rectified flow transformers for high-resolution image synthesis},
  author={Esser, Patrick and Kulal, Sumith and Blattmann, Andreas and Entezari, Rahim and M{\"u}ller, Jonas and Saini, Harry and Levi, Yam and Lorenz, Dominik and Sauer, Axel and Boesel, Frederic and others},
  booktitle={Forty-first international conference on machine learning},
  year={2024}
}

@article{heusel2017gans,
  title={Gans trained by a two time-scale update rule converge to a local nash equilibrium},
  author={Heusel, Martin and Ramsauer, Hubert and Unterthiner, Thomas and Nessler, Bernhard and Hochreiter, Sepp},
  journal={Advances in neural information processing systems},
  volume={30},
  year={2017}
}

@article{ho2020denoising,
  title={Denoising diffusion probabilistic models},
  author={Ho, Jonathan and Jain, Ajay and Abbeel, Pieter},
  journal={Advances in neural information processing systems},
  volume={33},
  pages={6840--6851},
  year={2020}
}

@article{loshchilov2017decoupled,
  title={Decoupled weight decay regularization},
  author={Loshchilov, Ilya and Hutter, Frank},
  journal={arXiv preprint arXiv:1711.05101},
  year={2017}
}

@inproceedings{yu2020bdd100k,
  title={Bdd100k: A diverse driving dataset for heterogeneous multitask learning},
  author={Yu, Fisher and Chen, Haofeng and Wang, Xin and Xian, Wenqi and Chen, Yingying and Liu, Fangchen and Madhavan, Vashisht and Darrell, Trevor},
  booktitle={Proceedings of the IEEE/CVF conference on computer vision and pattern recognition},
  pages={2636--2645},
  year={2020}
}

@inproceedings{zhu2016traffic,
  title={Traffic-sign detection and classification in the wild},
  author={Zhu, Zhe and Liang, Dun and Zhang, Songhai and Huang, Xiaolei and Li, Baoli and Hu, Shimin},
  booktitle={Proceedings of the IEEE conference on computer vision and pattern recognition},
  pages={2110--2118},
  year={2016}
}

@inproceedings{xiao2017joint,
  title={Joint detection and identification feature learning for person search},
  author={Xiao, Tong and Li, Shuang and Wang, Bochao and Lin, Liang and Wang, Xiaogang},
  booktitle={Proceedings of the IEEE conference on computer vision and pattern recognition},
  pages={3415--3424},
  year={2017}
}

@article{bai2023qwen,
  title={Qwen technical report},
  author={Bai, Jinze and Bai, Shuai and Chu, Yunfei and Cui, Zeyu and Dang, Kai and Deng, Xiaodong and Fan, Yang and Ge, Wenbin and Han, Yu and Huang, Fei and others},
  journal={arXiv preprint arXiv:2309.16609},
  year={2023}
}

@article{wu2025qwen,
  title={Qwen-image technical report},
  author={Wu, Chenfei and Li, Jiahao and Zhou, Jingren and Lin, Junyang and Gao, Kaiyuan and Yan, Kun and Yin, Sheng-ming and Bai, Shuai and Xu, Xiao and Chen, Yilei and others},
  journal={arXiv preprint arXiv:2508.02324},
  year={2025}
}

@article{hu2022lora,
  title={Lora: Low-rank adaptation of large language models.},
  author={Hu, Edward J and Shen, Yelong and Wallis, Phillip and Allen-Zhu, Zeyuan and Li, Yuanzhi and Wang, Shean and Wang, Lu and Chen, Weizhu and others},
  journal={ICLR},
  volume={1},
  number={2},
  pages={3},
  year={2022}
}
\bibliographystyle{spiebib} 

\end{document}